\documentclass[letterpaper]{article} 
\usepackage{aaai2026}  
\usepackage{times}  
\usepackage{helvet}  
\usepackage{courier}  
\usepackage[hyphens]{url}  
\usepackage{graphicx} 
\usepackage{natbib}  
\usepackage{caption} 
\usepackage{algorithm}
\usepackage{algorithmic}
\usepackage{amsmath}
\usepackage{booktabs}
\usepackage{xcolor} 

\usepackage{newfloat}
\usepackage{listings}
\DeclareCaptionStyle{ruled}{labelfont=normalfont,labelsep=colon,strut=off} 
\floatstyle{ruled}
\newfloat{listing}{tb}{lst}{}
\floatname{listing}{Listing}
\title{ML-EcoLyzer: Quantifying the Environmental Cost of Machine Learning Inference Across Frameworks and Hardware}
\author{
    Jose Marie Antonio Mi\~noza,
    Rex Gregor Laylo,
    Christian F Villarin,
    Sebastian C. Iba\~nez
}
\affiliations{
    Center for AI Research (CAIR) \\
    Department of Education, Philippines
}

\usepackage{bibentry}

\begin{document}

\maketitle

\begin{abstract}
Machine learning inference occurs at a massive scale, yet its environmental impact remains poorly quantified, especially on low-resource hardware. We present \textbf{ML-EcoLyzer}, a cross-framework tool for measuring the carbon, energy, thermal, and water costs of inference across CPUs, consumer GPUs, and datacenter accelerators. The tool supports both classical and modern models, applying adaptive monitoring and hardware-aware evaluation.

We introduce the \textit{Environmental Sustainability Score (ESS)}, which quantifies the number of effective parameters served per gram of CO$_2$ emitted. Our evaluation covers over 1,900 inference configurations, spanning diverse model architectures, task modalities (text, vision, audio, tabular), hardware types, and precision levels. These rigorous and reliable measurements demonstrate that quantization enhances ESS, huge accelerators can be inefficient for lightweight applications, and even small models may incur significant costs when implemented suboptimally. ML-EcoLyzer sets a standard for sustainability-conscious model selection and offers an extensive empirical evaluation of environmental costs during inference.
\end{abstract}


\section{Introduction}

Concerns about the environmental impact of artificial intelligence have grown as machine learning (ML) systems are increasingly deployed in a wide range of real-world applications, from recommender systems and search engines to conversational agents and edge devices. These models now run on both resource-constrained hardware and large-scale datacenters, creating new demands for sustainable and efficient deployment strategies.

While the carbon footprint of large-scale model training has been well studied~\cite{strubell2019energy, patterson2021carbon}, the environmental cost of inference remains underexplored—even though inference workloads often vastly outnumber training runs in practical settings~\cite{desislavov2023trends}. In production, inference is now the principal contributor to energy use and emissions in ML systems~\cite{patterson2022carbon}.

Despite growing awareness, there is no widely adopted standard for benchmarking the environmental impact of ML inference. Existing tools and benchmarks such as MLPerf~\cite{mattson2020mlperf} focus primarily on performance metrics (e.g., throughput, latency), and typically address only training or specific frameworks. Adaptive, cross-framework measurement protocols that cover both classical and modern models, diverse modalities, and heterogeneous hardware remain lacking.

\begin{figure}[t]
\centering
\includegraphics[width=0.9\columnwidth]{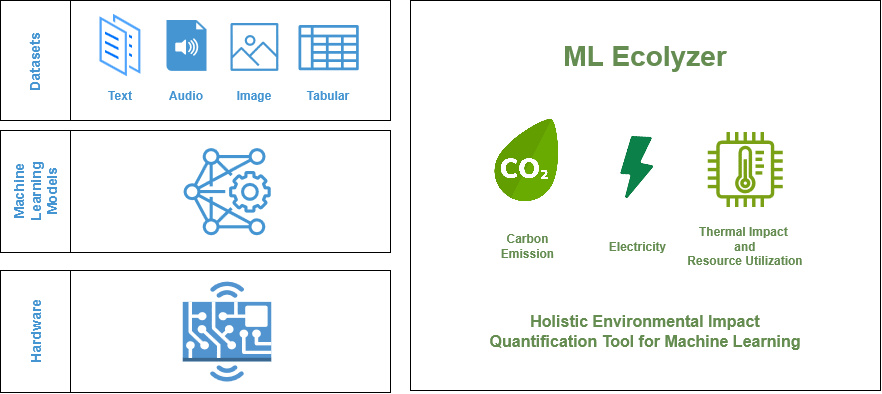}
\caption{Overview of ML-EcoLyzer. The framework quantifies the environmental impact of machine learning inference by integrating various dataset modalities, diverse model architectures, and hardware platforms.}
\label{fig:ml_ecolyzer_overview}
\end{figure}

To address this gap, we present \textbf{ML-EcoLyzer}, an open-source, extensible tool for evaluating the environmental impact of ML inference workloads (see Figure~\ref{fig:ml_ecolyzer_overview}). ML-EcoLyzer supports a wide array of tasks—including text, vision, audio, and tabular—and monitors real-time energy use, carbon emissions, thermal conditions, and water consumption across CPUs, consumer GPUs, and datacenter accelerators. Its adaptive monitoring engine enables consistent evaluation in both high-performance and low-resource environments. Throughout this work, we report environmental costs on a \textit{per-inference} basis, defined as the total environmental impact (energy, carbon, water) of processing a single input sample through the complete model pipeline—for example, one text prompt-response cycle, one image classification, or one audio transcription.

We also introduce the \textit{Environmental Sustainability Score (ESS)}, a new metric that quantifies the number of effective parameters served per gram of CO$_2$ emitted, enabling fair comparisons across models of different sizes, quantization levels, and operational footprints. Our empirical analysis, spanning over 1,900 inference configurations, shows that quantization substantially reduces carbon cost, and that task, model, and hardware choices all significantly affect environmental efficiency.

The main contributions of this work are:
\begin{itemize}
    \item We present \textbf{ML-EcoLyzer}, an open-source, framework-agnostic tool for evaluating the environmental impact of machine learning inference across both classical and modern architectures.
    \item We propose the \textit{Environmental Sustainability Score (ESS)}, a parameter-normalized metric for carbon cost comparison across models, hardware platforms, and quantization settings.
    \item We provide a comprehensive empirical assessment covering a wide range of modalities (text, vision, audio, tabular), model families, hardware tiers, and inference precisions.
\end{itemize}

This study provides practical tools and indicators to monitor sustainability inference time, enabling actionable recommendations for resource-constrained deployments when environmental efficiency is essential.

\section{Related Work}

\subsection{Carbon Footprint Analysis in Training Workloads}

Early investigations of the environmental impact of machine learning have focused primarily on model training. ~\cite{strubell2019energy} quantified the emissions of large-scale NLP training pipelines, estimating over 284,000 kg of CO$_2$ for BERT-large under hyperparameter tuning. ~\cite{patterson2021carbon} proposed a structured emission model incorporating energy consumption ($E$), grid carbon intensity ($CI$), and datacenter efficiency (PUE):

\begin{equation}
\text{CO}_2 = E \times CI \times \text{PUE}
\end{equation}

While foundational, this body of work emphasizes training cost and lacks resolution at the inference stage, where the majority of real-world energy expenditure occurs~\cite{strubell2019energy, patterson2021carbon, morrison2025holistic}.

\subsection{Environmental Reporting Tools and Protocols}

Several tools have emerged to track and report training emissions. CodeCarbon~\cite{codecarbon2022} and CarbonTracker~\cite{carbontracker2020} estimate carbon output using regional carbon intensity data and hardware power profiles. The Experiment Impact Tracker~\cite{henderson2020towards} provides a framework for transparent emissions reporting in academic settings. However, these tools generally lack support for inference workflows, operate within narrow framework boundaries, and do not address hardware thermal or water overheads.

ML-EcoLyzer extends these capabilities by supporting inference-time profiling across a wide range of model types and hardware tiers, including CPU-only and edge-oriented deployments.

\subsection{Lifecycle Benchmarks and Deployment Cost Modeling}

Standardized frameworks for evaluating the environmental impact of machine learning have primarily focused on training workloads. For example, the works of ~\cite{lacoste2019quantifying} and ~\cite{morrison2025holistic} provide systematic approaches for estimating and reporting carbon emissions, water usage, and energy consumption during model development and training. These methods represent important progress toward reproducible and transparent sustainability assessment.

However, inference workloads, which often account for the majority of operational ML activity, are not systematically benchmarked in these frameworks. Most available methodologies do not provide protocols for measuring environmental impact during the deployment phase, particularly across diverse tasks, modalities, and hardware types.

Our work addresses this limitation by introducing explicit, cross-framework, inference-time measurement protocols that cover a wide range of data modalities, model classes, and hardware configurations.

\subsection{Model Compression and Inference Optimization}

A variety of techniques have been proposed to improve inference efficiency, including knowledge distillation~\cite{hinton2015distilling}, pruning~\cite{han2015learning}, and quantization~\cite{jacob2018quantization, frantar2023gptq}. These methods reduce model size and compute requirements without significantly degrading accuracy. However, their impact on environmental metrics—such as carbon emissions and water usage—has been less systematically evaluated.

This study empirically benchmarks such techniques under consistent conditions to quantify their environmental benefits, particularly in constrained hardware scenarios.

\subsection{Thermal and Water Footprint Considerations}

Recent work has expanded the sustainability focus beyond carbon. ~\cite{dodge2022measuring} identified geographic variation in datacenter water and energy intensity, and  ~\cite{wu2022sustainable} called for broader environmental metrics beyond CO$_2$. Yet few tools incorporate these factors in a reusable benchmarking framework.

ML-EcoLyzer includes water usage estimation based on energy-to-water conversion coefficients and tracks thermal behavior in both CPU and GPU-bound workloads.

\subsection{Inference Benchmarking Gaps}

Significant progress has been made in standardizing the evaluation of machine learning models, especially for training efficiency and carbon emissions. For example, MLPerf~\cite{mattson2020mlperf} and DAWNBench~\cite{coleman2017dawnbench} provide industry-wide benchmarks for speed and accuracy, while ~\cite{lacoste2019quantifying} and ~\cite{morrison2025holistic} have introduced tools and methodologies for estimating and reporting the environmental impact of training, including holistic resource tracking.

Despite these advances, existing frameworks rarely address inference workloads in a systematic and cross-framework manner. Most prior work either focuses on training or is limited to carbon estimation without standardized protocol for measuring energy, water, or emissions during deployment. Furthermore, comprehensive benchmarking across diverse modalities (e.g., text, vision, audio, tabular), hardware platforms, and both classical and neural architectures remains largely unexplored for inference-time sustainability.

This work addresses these gaps by providing:
\begin{itemize}
    \item Framework-agnostic support for inference evaluation on both classical and neural models,
    \item Inclusion of multiple environmental dimensions (energy, emissions, thermal, water), and
    \item A standardized benchmarking protocol adaptable to varied hardware tiers and deployment scenarios.
\end{itemize}

Together, these contributions establish a foundation for principled, reproducible environmental analysis of inference workloads in real-world settings.

\section{ML Ecolyzer}

This section formalizes the metrics used in ML-EcoLyzer for assessing the environmental impact of machine learning inference. These include carbon emissions, energy consumption, water footprint, and the proposed Environmental Sustainability Score (ESS). All metrics are reported on a \textit{per-inference} basis—that is, the environmental cost of processing a single input sample through the complete model pipeline, enabling direct comparison across models and hardware configurations.

\subsection{Carbon Emissions Estimation}

Following established methodologies ~\cite{patterson2021carbon, codecarbon2022}, carbon emissions (in kg CO$_2$) are calculated using:

\begin{equation}
\text{CO}_2 \text{ (kg)} = E \text{ (kWh)} \times CI \text{ (kg CO}_2\text{/kWh)} \times \text{PUE}
\end{equation}

where:
\begin{itemize}
    \item $E$ is the total energy consumed during inference (in kWh),
    \item $CI$ is the carbon intensity of the regional power grid (in kg CO$_2$/kWh),
    \item $\text{PUE}$ denotes Power Usage Effectiveness (dimensionless), which accounts for infrastructure overhead.
\end{itemize}

For hardware classification, ML-EcoLyzer uses tier-specific PUE values based on industry benchmarks and datacenter efficiency studies~\cite{koomey2011growth, masanet2020recalibrating, uptime2022pue}: 1.1 for CPU-only systems (typical of edge and desktop environments with minimal cooling infrastructure), 1.2 for desktop-class GPUs (e.g., RTX, GTX series with moderate cooling requirements), and 1.4 for datacenter GPUs (e.g., A100, T4 reflecting enterprise cooling infrastructure overhead). These values align with reported data center efficiency ranges of 1.1-1.8 for modern facilities~\cite{uptime2022pue}.

\subsection{Energy Profiling and Power Monitoring}

Energy consumption (in kWh) is computed as the integral of instantaneous power over time:

\begin{equation}
E \text{ (kWh)} = \frac{1}{3600} \int_0^T P(t) \, dt
\end{equation}

where $P(t)$ is instantaneous power (in watts) at time $t$, and $T$ is the total duration of inference (in seconds). Power data is collected at adaptive sampling rates (typically 1–5 Hz) using system-level monitors (e.g., NVIDIA-SMI, \texttt{ psutil}) and validated power models.

\subsection{Water Footprint Estimation}
ML-EcoLyzer estimates water usage (in liters) based on real-time power monitoring, regional water intensity factors, and device-specific cooling overhead, following the methodology established in ~\cite{lacoste2019quantifying}:
\begin{equation}
\text{Water} = P_{\text{mon}} \times t \times WI_{\text{reg}} \times O_{\text{cool}} \times O_{\text{infra}}
\end{equation}
where all variables are in appropriate units: $P_{\text{mon}}$ (kW), $t$ (h), $WI_{\text{reg}}$ (L/kWh), and $O_{\text{cool}}$, $O_{\text{infra}}$ (dimensionless). Here, $P_{\text{mon}}$ represents power consumption estimated from real-time monitoring of CPU, GPU, and system utilization during ML workload execution, converted to energy consumption over time $t$. $WI_{\text{reg}}$ represents regional water intensity coefficients from the framework's comprehensive database covering 25+ regions, ranging from $1.2$~L/kWh (Iceland, geothermal/hydro) to $4.8$~L/kWh (Middle East, oil/gas generation), based on the calculator methodology~\cite{lacoste2019quantifying}. $O_{\text{cool}}$ represents device-specific cooling overhead factors ($1.0$× for low-power devices to $1.4$× for data centers) and $O_{\text{infra}}$ accounts for data center infrastructure water usage ($1.0$× to $1.2$×). The framework automatically detects regional context through locale, timezone, and cloud provider environment variables, applying validated coefficients with hardware-aware overhead calculations. This monitored-energy approach provides more accurate water footprint estimates than theoretical calculations, supporting comprehensive environmental impact assessment that extends beyond carbon emissions to include actual water resource consumption~\cite{wu2022sustainable,patterson2021carbon}.

\subsection{Effective Parameters}

To enable fair comparison across models of varying size and quantization, ML-EcoLyzer introduces the notion of \textit{Effective Parameters}. While raw parameter count reflects model capacity, it does not account for environmental differences due to quantization or precision. The effective parameter count linearly scales the total by the representational granularity:

\begin{equation}
\text{Effective Parameters (M)} = \frac{N \times QF}{10^6}
\end{equation}

where $N$ is the total number of parameters and $QF$ is the quantization factor, representing the bit-width scaling. The default $QF$ values are: (a) $1.0$ for FP32, (b) $0.5$ for FP16, and (c) $0.25$ for INT8, as supported by energy and throughput scaling in mixed-precision profiling~\cite{micikevicius2017mixed, jacob2018quantization}. In models using heterogeneous precision, $QF$ is computed as a weighted average across all layers. This linear scaling reflects the observation that lower-precision inference reduces memory, compute, and energy usage nearly proportionally, especially on hardware with native support for quantized operations~\cite{jacob2018quantization, frantar2023gptq}.

\subsection{Environmental Sustainability Score (ESS)}

The core metric proposed in this work is the Environmental Sustainability Score (ESS), defined as:

\begin{equation}
\text{ESS} = \frac{\text{Effective Parameters (M)}}{\text{CO}_2 \text{ (g)}}
\end{equation}

ESS measures how many effective parameters can be served per gram of CO$_2$ emitted. Higher ESS values indicate more sustainable inference configurations. The metric supports fair comparisons between full-precision and quantized models, as well as between traditional and modern architectures.

We chose parameter-based normalization over alternatives such as FLOPs-based or energy-per-token metrics for several reasons: (1) \textit{Hardware agnosticism}: FLOPs vary dramatically by hardware implementation and compiler optimization, making cross-platform comparisons unreliable, whereas effective parameters provide a consistent measure of model capacity; (2) \textit{Quantization awareness}: ESS inherently accounts for bit-width through the quantization factor ($QF$), enabling fair comparison across precision levels (FP32, FP16, INT8); (3) \textit{Multi-modal applicability}: Unlike token-based metrics limited to sequence models, parameter normalization generalizes across text, vision, audio, and tabular domains. ESS should always be interpreted alongside absolute emissions to avoid favoring large models with high per-parameter efficiency but unsustainable total footprints~\cite{desislavov2023trends}.

\subsection{Thermal Efficiency Considerations}

ML-EcoLyzer captures temperature traces during inference and flags thermally inefficient regimes, particularly when GPU temperature exceeds 80°C or CPU exceeds 85°C. Such regimes incur cooling overheads, which are estimated and added to the total energy budget. These thermal adjustments follow hardware datasheet modeling and ASHRAE thermal envelope guidance.

\subsection{Quantization Impact Tracking}

To assess the environmental benefit of quantization, the framework tracks baseline energy and water consumption during inference execution, then applies \textit{estimated quantization savings factors} to project multi-precision benefits. Water consumption estimates are calculated using regional water intensity factors and device-specific cooling overhead:


\begin{align}
W_{\text{baseline}}
  &= P_{\text{measured}} \times I_{\text{region}} \nonumber\\[-2pt]
  &\quad \times O_{\text{cooling}} \times O_{\text{infrastructure}}
  \label{eq:wbase}\\[4pt]
\text{\shortstack{Energy Savings (\%)\\(estimated)}}
  &= 100 \times
     \frac{
       E_{\text{measured}} - E_{\text{estimated\_quantized}}
     }{
       E_{\text{measured}}
     }
     \label{eq:esave}\\[4pt]
\text{\shortstack{Water Savings (\%)\\(estimated)}}
  &= 100 \times
     \frac{
       W_{\text{baseline}} - W_{\text{estimated\_quantized}}
     }{
       W_{\text{baseline}}
     }
     \label{eq:wsave}
\end{align}

where $P_{\text{measured}}$ is the measured power consumption, $I_{\text{region}}$ is the regional water intensity factor (1.2-4.8 L/kWh, \cite{codecarbon2022, lacoste2019quantifying, micikevicius2017mixed}), $O_{\text{cooling}}$ and $O_{\text{infrastructure}}$ are device-specific overhead factors, and the quantized variants are estimated using predetermined savings factors. These estimates provide rapid quantization impact assessment without requiring the computational overhead of actual multi-precision inference execution. The framework supports multiple precision modes (FP32, FP16, INT8) through its quantization configuration system, enabling the potential implementation of empirical measurements. Water equivalents are provided in practical units (bottles saved, gallons conserved) alongside comprehensive regional water footprint analysis.

\section{Experiments}

We present a comprehensive evaluation of inference-time environmental costs, covering over 1,900 configurations with ML-EcoLyzer. Our study includes transformer-based LLMs, classical models, and a range of task modalities, with all results normalized to single-sample inference. This section explores how emissions, water use, and ESS are shaped by model architecture, hardware platform, task, and precision.

\begin{table*}[t]
\centering
\begin{tabular}{lccc}
\toprule
\textbf{Model Family} & \textbf{CO$_2$ (kg, $\mu\pm\sigma$)} & \textbf{Water (L, $\mu\pm\sigma$)} & \textbf{ESS (MP/g, $\mu\pm\sigma$)} \\
\midrule
GPT         & 0.121~$\pm$~0.317 & 1.28~$\pm$~3.43 & 882~$\pm$~4638 \\
OPT         & 0.098~$\pm$~0.255 & 1.05~$\pm$~2.73 & 486~$\pm$~329 \\
Qwen 2      & 0.016~$\pm$~0.009 & 0.14~$\pm$~0.06 & 1233~$\pm$~10,968 \\
Phi 3       & 0.015~$\pm$~0.002 & 0.11~$\pm$~0.01 & 585~$\pm$~317 \\
OLMo        & 0.020~$\pm$~0.010 & 0.21~$\pm$~0.11 & 237~$\pm$~162 \\
Gemma 2     & 0.038~$\pm$~0.088 & 0.18~$\pm$~0.66 & 953~$\pm$~5749 \\
LLaMA 2     & 0.029~$\pm$~0.066 & 0.32~$\pm$~0.71 & 678~$\pm$~252 \\
Playground  & 0.016~$\pm$~0.001 & 0.17~$\pm$~0.01 & 158~$\pm$~12 \\
HuBERT      & 0.0057~$\pm$~0.0060 & 0.061~$\pm$~0.062 & 108~$\pm$~65 \\
Whisper     & 0.0062~$\pm$~0.0052 & 0.064~$\pm$~0.053 & 96~$\pm$~108 \\
\bottomrule
\end{tabular}
\caption{Environmental cost by model family (per-inference basis). Metrics show average CO$_2$ emissions (kg), water consumption (L), and Environmental Sustainability Score (ESS: million effective parameters per gram CO$_2$) for processing a single input sample. Mean~$\pm$~std aggregated across model variants (e.g., Qwen 1.8B, 4B, 7B) and hardware configurations. High standard deviations reflect substantial variance in hardware utilization, model sizes within families, and deployment configurations.}
\label{tab:family-results}
\end{table*}

\paragraph{Model Family Analysis.}
Environmental impact varies by model family (Table~\ref{tab:family-results}). Legacy architectures such as GPT and OPT are among the most emission-intensive, with average emissions per inference of 0.121~$\pm$~0.317~kg and 0.098~$\pm$~0.255~kg CO$_2$, respectively. In contrast, newer models such as Qwen 2 (0.016~$\pm$~0.009~kg) and Phi 3 (0.015~$\pm$~0.002~kg) demonstrate much greater efficiency. These modern families also achieve substantially higher ESS, sometimes by an order of magnitude. This finding runs counter to the common assumption that larger or newer models are always more environmentally costly; our empirical results, as well as recent literature~\cite{desislavov2023trends}, demonstrate that architectural and precision advances can significantly reduce per-parameter emissions.

\begin{figure}[t]
\centering
\includegraphics[width=0.9\columnwidth]{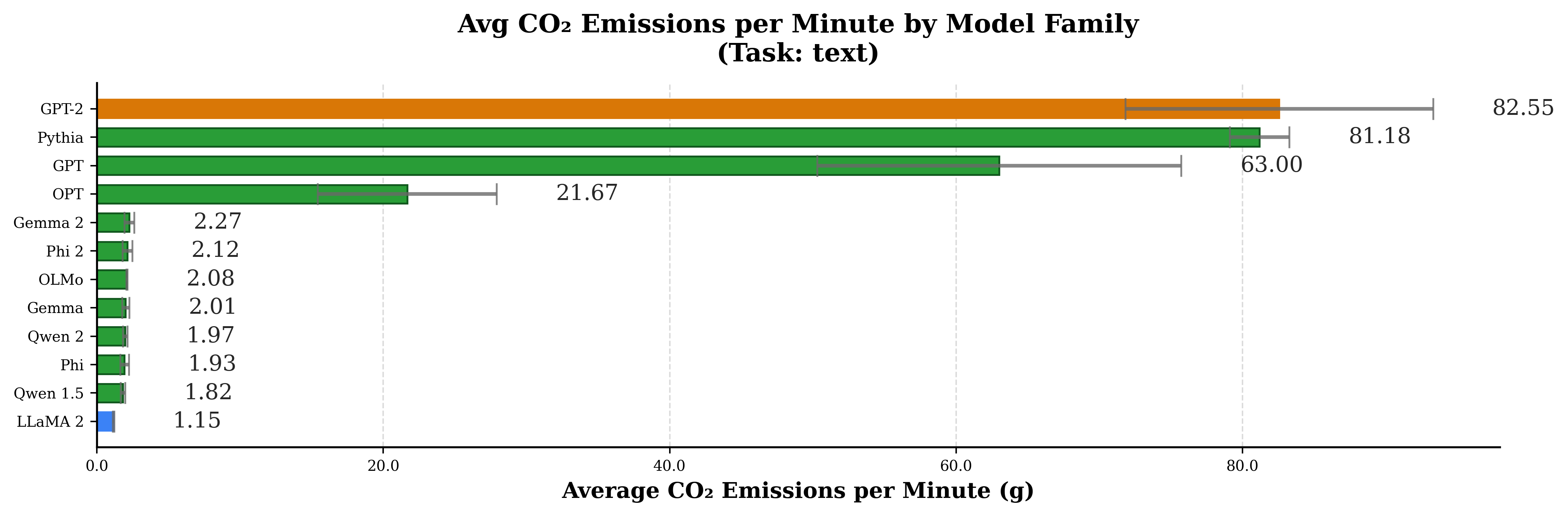}
\caption{Average CO$_2$ emissions per minute by model family on text generation tasks (rate-based metric). This complements Table~\ref{tab:family-results}, which reports per-inference costs. Per-minute metrics reveal throughput efficiency during sustained operation, while per-inference metrics show task-level environmental cost. Error bars: standard deviation.}
\label{fig:co2_text_family}
\end{figure}

Figure~\ref{fig:co2_text_family} illustrates the dramatic environmental gap between model generations for text workloads. Despite similar parameter counts, recent releases like Qwen 2, OLMo, and LLaMA 2 deliver much lower per-minute emissions than earlier models such as GPT-2 or Pythia. This shows that relying on older, smaller models can be environmentally counterproductive; up-to-date architectures yield larger sustainability gains than downsizing alone.

\paragraph{Hardware Platform Analysis.}
Table~\ref{tab:device-results} shows that the environmental cost of inference is highly sensitive to hardware choice. Datacenter GPUs such as the A100 deliver low average emissions (0.024~$\pm$~0.053~kg CO$_2$ per run) and the highest mean ESS, but only when heavily utilized~\cite{mattson2020mlperf}. Suboptimal batching or small workloads can reduce efficiency, even on advanced accelerators, as supported by our experimental results and by findings in large-scale ML inference benchmarks~\cite{mattson2020mlperf}.

\begin{table*}[t]
\centering
\begin{tabular}{lccc}
\toprule
\textbf{Device Category} & \textbf{CO$_2$ (kg, $\mu\pm\sigma$)} & \textbf{Water (L, $\mu\pm\sigma$)} & \textbf{ESS (MP/g, $\mu\pm\sigma$)} \\
\midrule
Cloud (Tesla T4)    & 0.097~$\pm$~0.238 & 0.34~$\pm$~0.40 & 991~$\pm$~8,538 \\
Datacenter (A100)   & 0.024~$\pm$~0.053 & 0.02~$\pm$~0.00 & 1,313~$\pm$~9,128 \\
Consumer (RTX 4090) & 0.019~$\pm$~0.025 & 0.003~$\pm$~0.003 & 190~$\pm$~266 \\
Consumer (GTX 1650) & 0.047~$\pm$~0.077 & 0.012~$\pm$~0.008 & 164~$\pm$~724 \\
CPU (sklearn)       & 0.052~$\pm$~0.078 & 0.031~$\pm$~0.029 & 0.09~$\pm$~0.47 \\
\bottomrule
\end{tabular}
\caption{Environmental cost by hardware category. ESS: MP/g (mean~$\pm$~std).}
\label{tab:device-results}
\end{table*}

Maximizing hardware utilization through batching or workload aggregation is essential for sustainability. Deploying LLMs on large accelerators delivers high ESS only if those devices are fully used; otherwise, smaller, well-matched hardware may be more sustainable.

\paragraph{Task Modality Analysis.}
Task type also affects environmental cost. Table~\ref{tab:task-results} shows that text generation (LLMs) has the highest absolute CO$_2$ per run, but, due to large parameter counts, still achieves high ESS. In contrast, classification and regression tasks, often dominated by traditional models implemented with scikit-learn and executed on the CPU, show extremely low ESS. This inefficiency arises because, as our measurements show (see Table~\ref{tab:task-results}), for small models running on general-purpose CPUs, fixed system overheads and idle resource consumption outweigh the actual computational work of inference. This is consistent with observations in recent system-level analyses~\cite{desislavov2023trends, barroso2007energy}. As a result, even “lightweight” pipelines can be less sustainable when deployed on CPUs without batching or hardware matching. This finding is significant for edge and resource-constrained deployments, where relying on classical ML does not necessarily translate to greener inference.

\begin{table*}[t]
\centering
\begin{tabular}{lccc}
\toprule
\textbf{Task Type} & \textbf{CO$_2$ (kg, $\mu\pm\sigma$)} & \textbf{Water (L, $\mu\pm\sigma$)} & \textbf{ESS (MP/g, $\mu\pm\sigma$)} \\
\midrule
Text Generation    & 0.134~$\pm$~0.269  & 0.30~$\pm$~0.60  & 2,397~$\pm$~13,400 \\
Classification     & 0.028~$\pm$~0.042  & 0.03~$\pm$~0.03  & 1.88~$\pm$~7.72 \\
Regression         & 0.074~$\pm$~0.089  & 0.07~$\pm$~0.09  & 1.18~$\pm$~6.63 \\
Image Generation   & 0.019~$\pm$~0.006  & 0.02~$\pm$~0.01  & 91.7~$\pm$~52.4 \\
Audio Processing   & 0.0060~$\pm$~0.0055 & 0.006~$\pm$~0.006 & 76.3~$\pm$~84.9 \\
Image Processing   & 0.0021~$\pm$~0.0005 & 0.002~$\pm$~0.001 & 35.2~$\pm$~16.3 \\
\bottomrule
\end{tabular}
\caption{Environmental cost by task type (per-inference basis). Task definitions: Text Generation (autoregressive LLMs: GPT, LLaMA, Qwen), Classification (image/text classification using ResNet, ViT, BERT), Regression (tabular prediction using scikit-learn models), Image Generation (diffusion and transformer-based synthesis: Stable Diffusion), Audio Processing (ASR and embedding: Whisper, HuBERT), Image Processing (classification and embedding extraction: DeiT, ResNet). ESS: million effective parameters per gram CO$_2$ (mean~$\pm$~std).}
\label{tab:task-results}
\end{table*}

\begin{figure}[t]
\centering
\includegraphics[width=0.9\columnwidth]{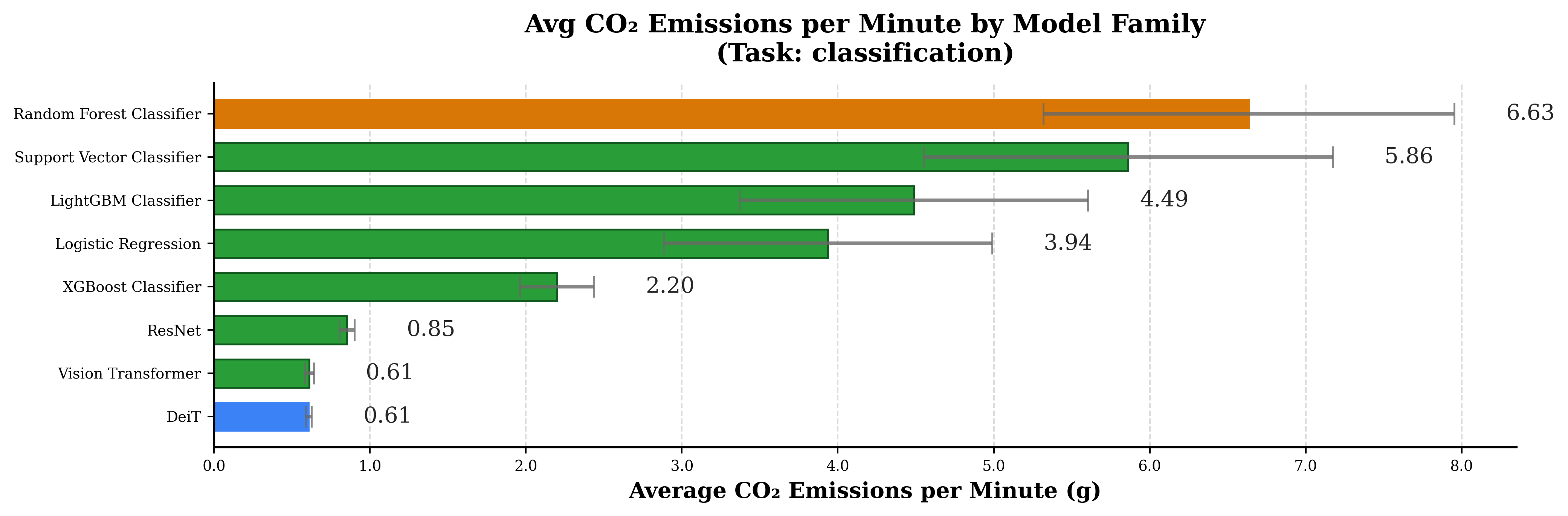}
\caption{Average CO$_2$ emissions per minute by model family on classification tasks. Error bars: standard deviation.}
\label{fig:co2_classification_family}
\end{figure}

Figure~\ref{fig:co2_classification_family} shows that ensemble methods like Random Forest and LightGBM are among the least efficient, often exceeding both linear and deep learning methods in per-minute emissions. Vision transformers and convolutional models, by contrast, offer lower emissions and higher throughput, especially when deployed on compatible hardware.

\paragraph{Quantization and Precision Effects.}
Many of the highest-ESS models are natively quantized or use reduced precision. As shown in Table~\ref{tab:quantization-results}, FP16 and INT8 models retain high accuracy (98.5\% and 94.2\%, respectively) while reducing power draw by 25\% to 55\% and water usage by over 0.01~L per inference. These empirical trends align with prior literature~\cite{jacob2018quantization,frantar2023gptq}, confirming quantization as a dominant lever for sustainable deployment.

\begin{table*}[t]
\centering
\begin{tabular}{lccc}
\toprule
\textbf{Precision} & \textbf{Power Savings} & \textbf{Accuracy Retention} & \textbf{Water Reduction (L)} \\
\midrule
FP32 & ---     & 100\%  & ---    \\
FP16 & 25\%    & 98.5\% & 0.0147  \\
INT8 & 55\%    & 94.2\% & 0.0152  \\
INT4 & 75\%    & 87.8\% & ---    \\
\bottomrule
\end{tabular}
\caption{Quantization benefits: power, energy, water, and accuracy (empirical and literature-supported).}
\label{tab:quantization-results}
\end{table*}

Most quantization gains are realized with minimal loss in accuracy for common transformer architectures. Sustainability benchmarks should always report model precision and leverage available quantized releases where possible.

\paragraph{Monitoring Sensitivity.}
Measurement protocol directly affects reported emissions, especially for short or bursty inference tasks. Table~\ref{tab:freq-results} shows that using a 1~Hz sampling rate can overestimate emissions by nearly 6\% compared to a 5~Hz standard, especially on fast models or accelerators. Adaptive monitoring strategies are necessary to ensure fair comparisons and avoid misleading results.

\begin{table}[htbp]
\centering
\begin{tabular}{lcc}
\toprule
\textbf{Sampling Rate (Hz)} & \textbf{Avg CO$_2$ (g)} & \textbf{Relative Error (\%)} \\
\midrule
1   & 27 & +5.8 \\
2   & 26 & +1.9 \\
5   & 25.5 & --- \\
\bottomrule
\end{tabular}
\caption{Effect of Sampling Frequency on CO$_2$ Estimation.}
\label{tab:freq-results}
\end{table}

In general, our results highlight that sustainable ML inference depends on architectural choices, hardware task alignment, precision, and measurement practices. In the next section, we discuss the implications and recommendations.

\section{Discussion}

This study provides a systematic quantification of the environmental impact of ML inference and establishes formal benchmarks for sustainability across model families, hardware, and deployment modalities. Our results demonstrate that, when emissions are normalized per effective parameter, recent LLMs often achieve higher Environmental Sustainability Scores (ESS) than traditional ML models. This effect is driven by the ability of transformer models to amortize energy costs over a large parameter space and to utilize hardware optimized for high-throughput inference.

\begin{figure*}[t]
\centering
\includegraphics[width=0.65\textwidth]{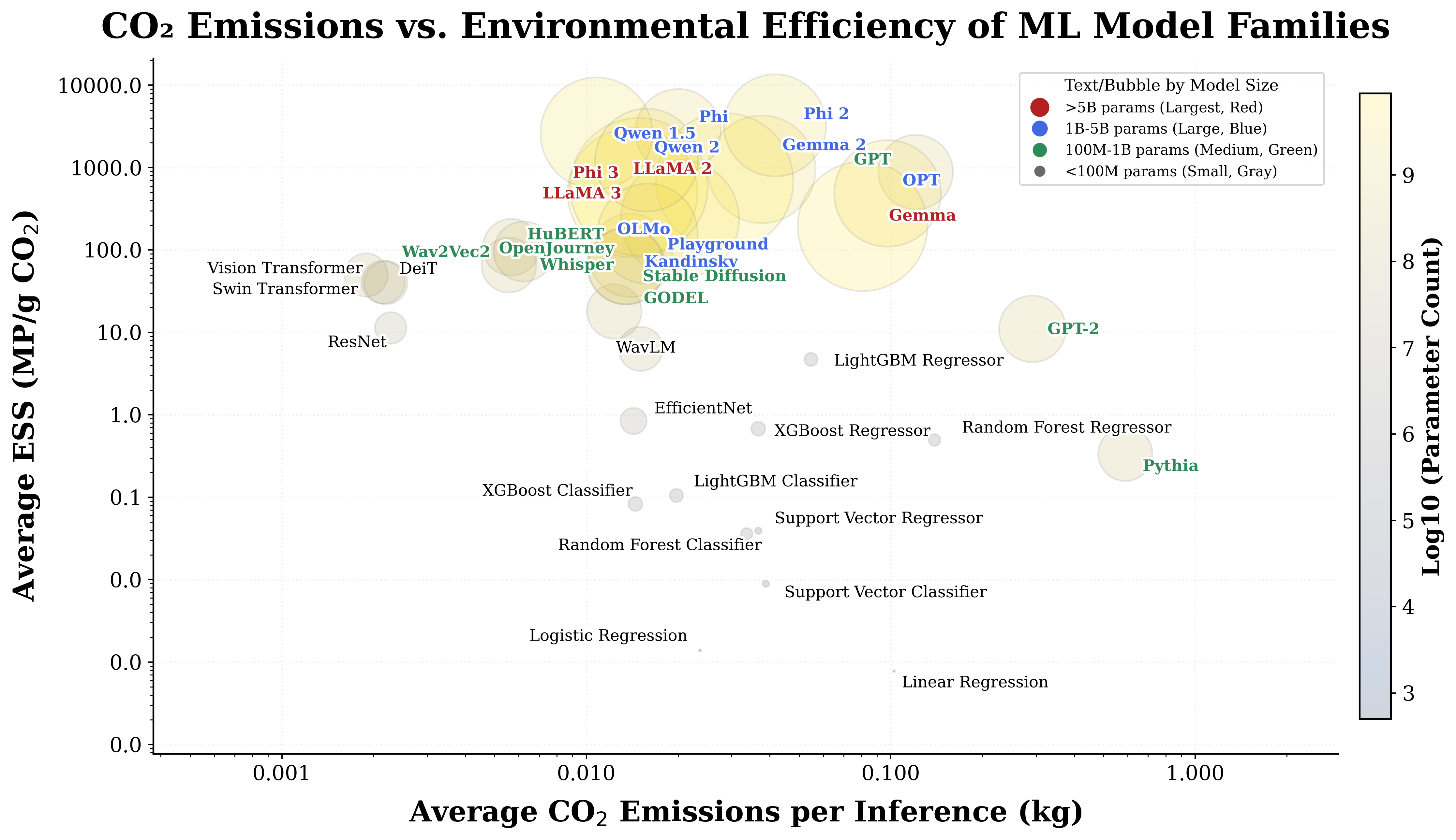}
\caption{Sustainability Score (ESS, MP/g CO$_2$) vs. CO$_2$ emissions per inference (kg) for all model families.}
\label{fig:ess_vs_co2_frontier}
\end{figure*}

A key empirical finding is that traditional models, such as those implemented in scikit-learn, typically run on CPUs, where fixed system overhead and low utilization lead to very low ESS, especially in single-inference or non-batched settings. Although these classical models emit less CO$_2$ per run in absolute terms, their sustainability per parameter is substantially lower than that of quantized transformer models deployed on GPUs or datacenter accelerators. Hardware alignment and reduced precision significantly boost ESS for modern models.

Figure~\ref{fig:ess_vs_co2_frontier} summarizes the efficiency landscape for all evaluated model families. The highest efficiency region, which combines low CO$_2$ emissions with high ESS, is dominated by modern transformer-based models, such as the Phi, Qwen, and LLaMA series, particularly in quantized form. Classical ML models, including Random Forest, Logistic Regression, and LightGBM, consistently exhibit low ESS despite small absolute emissions. This efficiency frontier makes clear that architectural advances, hardware-software co-design, and quantization provide greater sustainability benefits than simply reducing model size.

Quantization consistently improves environmental efficiency. INT8 and FP16 models reduce per-inference emissions substantially, with negligible accuracy loss in most transformer-based tasks. These results confirm the impact of quantized deployments and support the release and adoption of lower-precision model variants.

Hardware configuration is a decisive factor in sustainability. Datacenter GPUs like the A100 deliver high ESS only under full utilization. For lightweight or edge workloads, consumer accelerators or CPUs are often more efficient. Careful alignment of workload, model, and hardware is essential for sustainable deployment.

Measurement protocol also affects results: sampling frequency and monitoring granularity have a measurable effect on emissions estimates, especially for brief or bursty inference tasks. Adaptive monitoring, as implemented in ML-EcoLyzer, ensures accuracy without introducing excess overhead.

This study establishes reliable empirical benchmarks for inference-time environmental impact and demonstrates that ESS serves as a robust metric for sustainability-aware machine learning deployment.  It is essential that both absolute and parameter-normalized sustainability metrics, including ESS, be standardized in benchmarking and deployment studies.

\section{Conclusion}

This paper presents ML-EcoLyzer, an extensible, cross-framework, and open-source package that is designed to measure the environmental impact of machine learning inference across model types, tasks, and hardware configurations. This study emphasizes the importance of inference, a frequently overlooked but critical phase when implementing machine learning systems, underscoring the environmental costs that extend beyond emissions during the training phase.

We propose the Environmental Sustainability Score (ESS), a quantization-aware metric that captures per-parameter emissions and enables fair efficiency comparisons across models of varying precision and scale. Through over 1,900 inference runs across four hardware tiers and multiple task modalities, our benchmark provides concrete evidence for the efficiency gains of quantization, the value of hardware-utilization matching, and the role of adaptive monitoring in sustainability-aware evaluation.

ML-EcoLyzer formalizes established trends and uncovers overlooked dynamics, including the inefficiency of conventional ML models operating on idle hardware and the measurement bias that arises from coarse sampling. Releasing the tool as open-source software fosters additional experimentation and the establishment of environmental benchmarks, especially in resource-constrained or low-latency deployment scenarios. This study offers a comprehensive framework in quantifying sustainability as a core goal in the design and assessment of machine learning systems.

Potential directions for future work include expanding ML-EcoLyzer to support batched and streaming inference scenarios, adding native integration for additional ML frameworks and hardware platforms, and incorporating real-time regional grid carbon intensity data for dynamic emissions estimation. We also see value in developing more granular, task-aware sustainability metrics and in extending the analysis to encompass training workloads and multi-model serving deployments. Finally, collaborating with industry partners could help validate and refine these tools in real-world, production-scale environments.

\bibliography{aaai2026}

\end{document}